\documentclass[suppldata]{interact}

\usepackage{epstopdf}% To incorporate .eps illustrations using PDFLaTeX, etc.
\usepackage{subfigure}% Support for small, `sub' figures and tables

\usepackage{natbib}% Citation support using natbib.sty
\bibpunct[, ]{(}{)}{;}{a}{}{,}% Citation support using natbib.sty
\renewcommand\bibfont{\fontsize{10}{12}\selectfont}% Bibliography support using natbib.sty

\usepackage{multirow}
\usepackage{makecell}
\usepackage{pgfplots}
\pgfplotsset{width=7cm,compat=1.13}

\theoremstyle{plain}% Theorem-like structures provided by amsthm.sty

\theoremstyle{definition}

\theoremstyle{remark}

\begin{document}

%\articletype{ARTICLE TEMPLATE}

\title{Semantically-consistent Landsat 8 image to Sentinel-2 image translation for alpine areas}

\author{
\name{
Mikhail Sokolov\textsuperscript{a},
Joni L. Storie\textsuperscript{b},
Christopher J. Henry\textsuperscript{a},
Christopher D. Storie\textsuperscript{b},
Jami Cameron\textsuperscript{b},
Rune Strand Ødegård\textsuperscript{c},
Vilma Zubinaite\textsuperscript{c},
Sverre Stikbakke\textsuperscript{c}
}
\affil{
\textsuperscript{a}Department of Applied Computer Science, University of Winnipeg, 515 Portage
Ave, Winnipeg, R3B 2E9, Manitoba, Canada;
\textsuperscript{b}Department of Geography, University of Winnipeg, 515 Portage Ave, Winnipeg, R3B 2E9, Manitoba, Canada;
\textsuperscript{c}Department of Manufacturing and Civil Engineering, Norwegian University of Science \& Technology (NTNU) Gjøvik, Postbox 191, NO-2802 Gjøvik, Norway}
}

\maketitle

\begin{abstract}
The availability of frequent and cost-free satellite images is in growing demand in the research world. Such satellite constellations as Landsat 8 and Sentinel-2 provide a massive amount of valuable data daily. However, the discrepancy in the sensors’ characteristics of these satellites makes it senseless to use a segmentation model trained on either dataset and applied to another, which is why domain adaptation techniques have recently become an active research area in remote sensing. In this paper, an experiment of domain adaptation through style-transferring is conducted using the HRSemI2I model to narrow the sensor discrepancy between Landsat 8 and Sentinel-2. This paper's main contribution is analyzing the expediency of that approach by comparing the results of segmentation using domain-adapted images with those without adaptation. The HRSemI2I model, adjusted to work with 6-band imagery, shows significant intersection-over-union performance improvement for both mean and per class metrics. A second contribution is providing different schemes of generalization between two label schemes - NALCMS 2015 and CORINE. The first scheme is standardization through higher-level land cover classes, and the second is through harmonization validation in the field.
\end{abstract}


\section{Introduction}

Semantic segmentation of satellite images using deep learning (DL) algorithms is becoming an important tool that remote sensing scientists use for land use land cover (LULC) map generation \citep{storie,  henry, alhs2020, advkl}. These DL techniques are relatively inexpensive, rapid to generate, and highly accurate. However, the variability of data from different satellite sensors poses a challenge for automation using segmentation models; these models are trained on a specific domain with specific characteristics (i.e., data, land cover types), and the inferences of land cover in another domain usually leads to low or poor accuracy. In recent research, \citet{sokolov} proposed a high-resolution semantically-consistent image-to-image translation model (HRSemI2I) to mitigate the discrepancy between domains with different spectral characteristics (i.e., sensors) while preserving semantic consistency. In this work, they transferred the style of the 4-band imagery acquired by the SPOT-6 satellite (the domain without semantic labels, called the {\em target} domain) to imagery obtained with WorldView-2 (the domain with available semantic labels, called the {\em source} domain). The resulting source images with the style of the target domain were then used for semantic segmentation model training, and the resulting segmentation model accuracy showed increases of $\sim$3\% - 10\% compared to the model without adaptation.

In this work, we consider using the same domain adaptation framework, however, it is now applied to imagery acquired using Landsat 8 (source domain) and Sentinel-2 (target domain). Landsat and Sentinel satellites are from the two most common sensors used for land cover mapping, since the data they produce are publicly available and easily accessible. Considering semantic consistency between these two data sources increases the average global revisit interval to approximately three days, which allows for more frequent surface monitoring in many regions at a medium spatial resolution \citep{chaves}. Thus, a method of domain adaptation is highly relevant to the remote sensing community.

In addition to incorporating data from the two most common satellites used for LULC mapping, this work explores the adaptation performance of the model proposed by \citet{sokolov} (expanded to 6-banded imagery) as well as using standardized and harmonized validation techniques to assess the domain adaptation map product.
The model was trained and validated on 8 general land cover classes (water, urban, forest, agriculture, grassland, wetland, snow and ice, barren). The LULC classes for British Columbia were obtained by generalizing the NALCMS 2015 land cover classes while the Norway classes were validated using generalization of the CORINE land use classes. This allowed for standardized validation of the model which was based on 100 random points to calculate Intersection over Union (IoU), also known as {\em Jaccard index} \citep{jaccard, iou}, for overall and per-class accuracy. Although the LULC detail is lost in this standardized validation method, it incorporated sample randomness and representation of classes required for validation statistics.

This standardized validation method was followed with harmonization validation between the 19 land cover classes in the NALCMS and 44 land use classes in Norway from the CORINE (31 out of 44 classes were represented for the Norway region of interest). When matching different LULC schema, standardization is the most common method used \citep{sohl2016, fonte2017, li2021}, however, a combination of both standardization and harmonization methods is shown to be more accurate \citep{li2021, khaldi2022}. Standardization validation is the practice of combining LULC classes into a general classification schema, where harmonization validation does direct comparison between schema. When reviewing these methods, no study was found to use field data for additional validation of a UDA model to aid in matching different LULC schema. This harmonization validation method was completed in the field near Raubergstulen in Oppland Municipality, Norway. The data collection for harmonized validation was opportunistic, e.g., based on accessibility of locations over a two week period, and collected information for NALCMS, CORINE and general LULC schema. The purpose of the field data was two fold. First, it allowed for better understanding of the errors in the domain adaptation model output (i.e., poor labels or incorrectly classifying land cover). Second, to assess correspondence between the LULC schema (i.e., needle leaf with coniferous forests with forests) to identify more than eight general land cover labels that could improve the model performance in future.

%This standardized validation method was followed with harmonization validation between the 19 land cover classes in the NALCMS and 44 land use classes in Norway from the CORINE (31 out of 44 classes were represented for the Norway region of interest). This harmonization validation method was completed in the field near Raubergstulen in Oppland Municipality, Norway. The data collection for harmonized validation was opportunistic, e.g., based on accessibility of locations over a two week period, and collected information for NALCMS, CORINE and general LULC schema. The purpose of the field data was two fold. First, it allowed for better understanding of the errors in the domain adaptation model output (i.e., poor labels or incorrectly classifying land cover). Second, to assess correspondence between the LULC schema (i.e., needle leaf with coniferous forests with forests) to identify more than eight general land cover labels that could improve the model performance in future.

\section{Methodology}
A complete description of the adaptation model can be found in \citet{sokolov}. Briefly, the results in this article are generated by the high-resolution semantically-consistent image-to-image translation (HRSemI2I) model, which employs the AdaIN \citep{adain} layer and aims to transfer the target domain’s style to the source images thereby preserving semantic consistency and per-pixel image
quality. The domain adaptation problem is solved through adversarial learning process using a style-transferring model. After this, a segmentation model is trained using original source samples and style-transferred source samples. Then, the target domain labels are generated and evaluated.
The style-transferring model is aimed to transfer style of the target domain (Sentinel-2) to the source domain (Landsat 8) samples.  Let us denote $S$ and $T$ as source domain and target domain, respectively. Then $f_S$ ($f_T$) represent a target (source) sample transformed to the style of the source (target) domain. The neural network that performs a style-transferring task consists of two parts. The first one is called a generator ($G$) which generates outputs that are indistinguishable from the real data. The second part is a discriminator ($D$), and its goal is to decide how close the data transformed by the generator sample is to the target distribution. Together, the generator and discriminator are being optimized to minimize the following loss function:
\begin{equation}\label{eq:gan_loss}
L_{GAN}(D,G)=\mathbb{E}[\log D(x_T)]+\mathbb{E}[\log (1-D(G(x_S)))],
\end{equation}
where $x_S$ and $x_T$ are samples from the source and target domains, respectively. There are two pairs of a generator and a discriminator, and each pair is for the direction of style transformation: source-to-target and target-to-source.

\begin{figure}
	\centering
	\includegraphics[width=12cm]{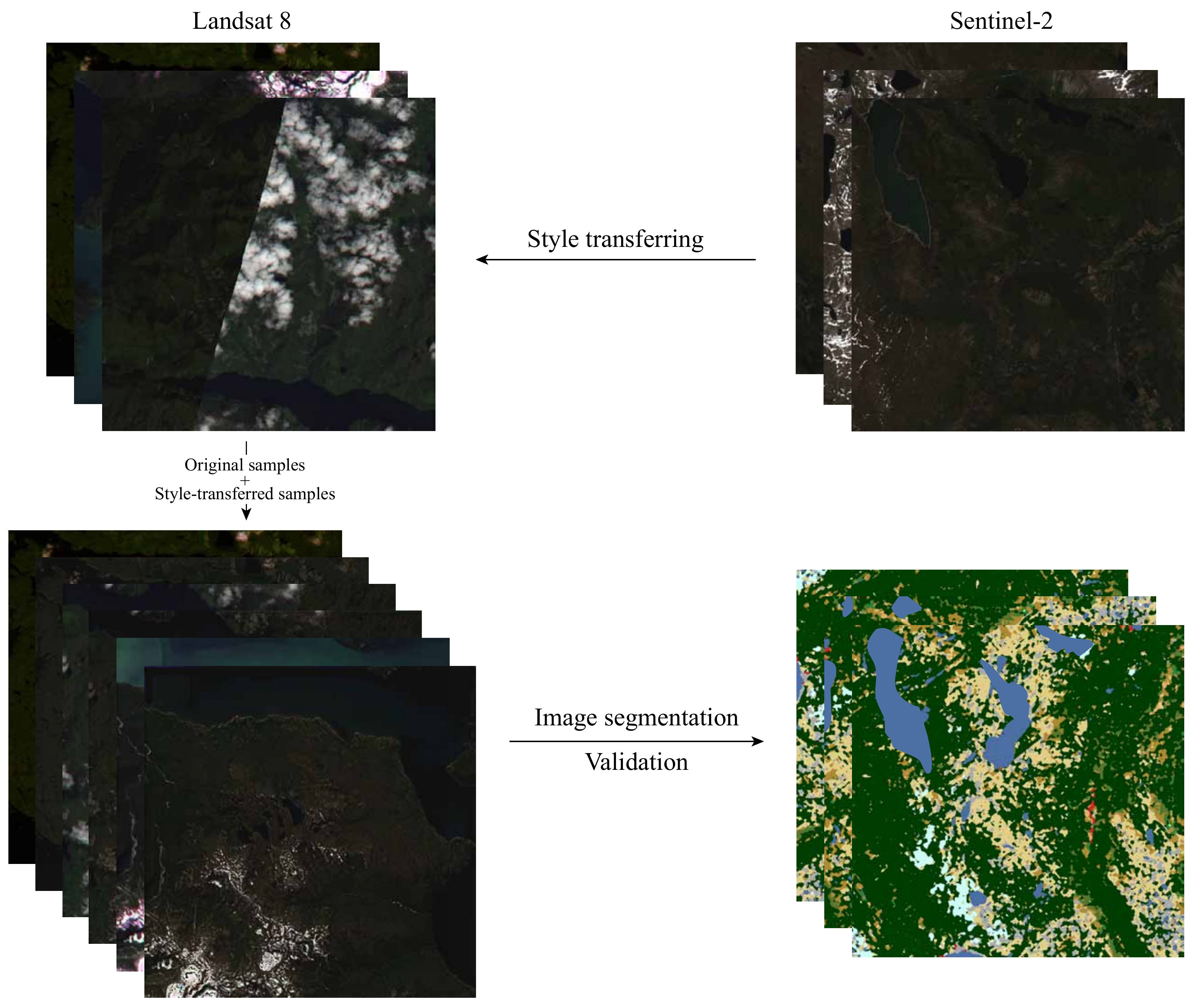}
	\caption{The proposed method. The style-transferring model transfer the style of the target domain to the source domain images. After this, the mixed dataset (consisted of original source and stylized source images) is used for the segmentation model training. The trained model is then validated on the target domain.}
	\label{fig:method}
\end{figure}

The segmentation model, which is used after the style-transferring part, is represented by the DeepLab v3 framework \citep{dl3}  with a modified number of input channels, in this case, equal to 6 bands of imagery. We use both original Landsat 8 samples and their stylized versions as a mixed dataset since it showed higher performance compared to using only stylized samples. The corresponding labels for the original samples and stylized are duplicated. After the segmentation model is trained, the validation set consists of the entire Sentinel-2 (target) dataset. The overall method is depicted on Figure~\ref{fig:method}.

\section{Experiments}
\begin{figure}
\centering
\subfigure[The source dataset region of interest (Canada).\label{fig:roi:bc}]{%
\resizebox*{7cm}{!}{\includegraphics{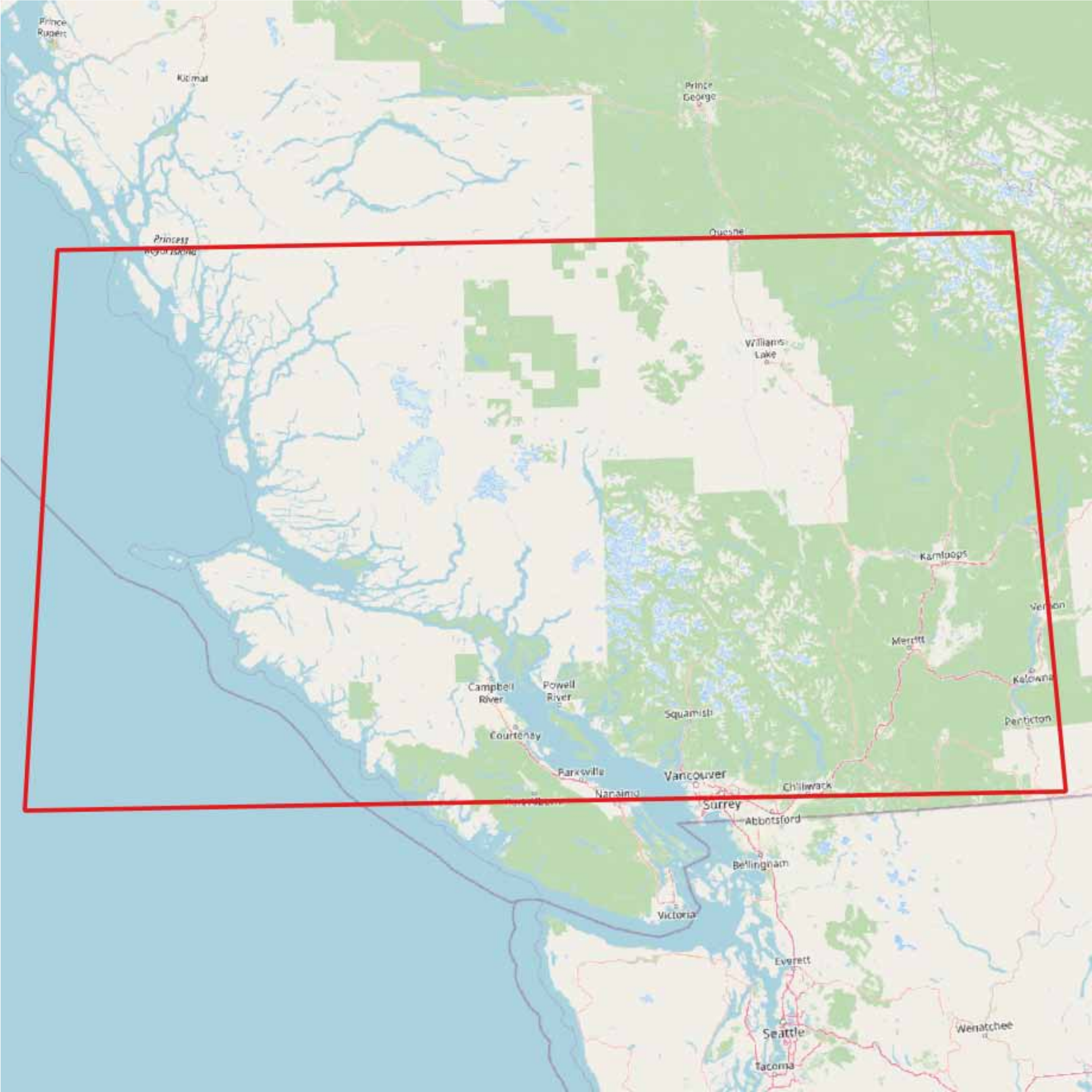}}}\hspace{5pt}
\subfigure[The target dataset region of interest (Norway).\label{fig:roi:nw}]{%
\resizebox*{7cm}{!}{\includegraphics{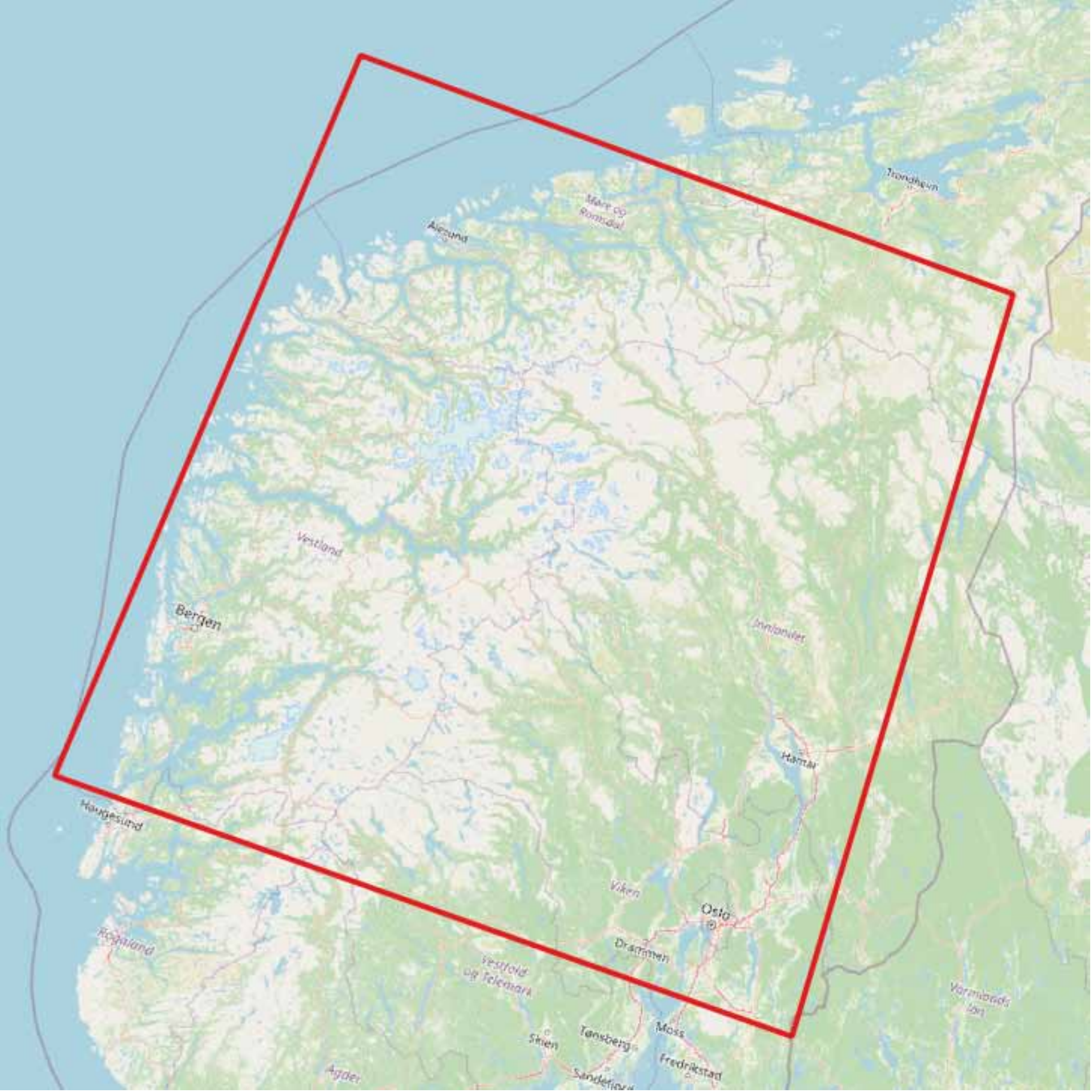}}}
\caption{The regions of interest for the source (a) and target (b) datasets.}
\end{figure}

\begin{figure}
\centering
\subfigure[Original pixel value distribution of the Landsat 8 dataset.\label{fig:distr:bafore}]{%
\resizebox*{7cm}{!}{\includegraphics{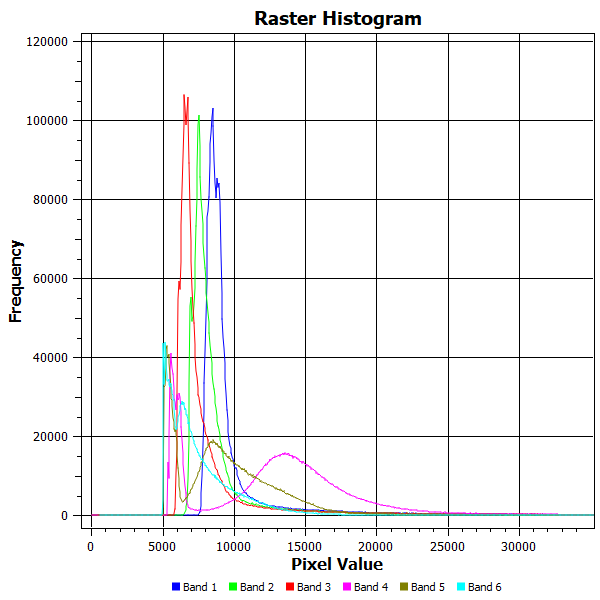}}}\hspace{5pt}
\subfigure[Shifted pixel value distribution of the Landsat 8 dataset.\label{fig:distr:after}]{%
\resizebox*{7cm}{!}{\includegraphics{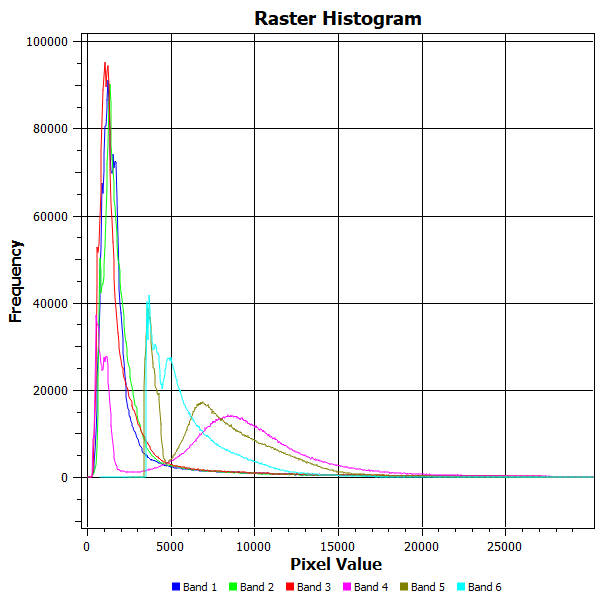}}}
\caption{The Landsat 8 dataset pixel value distribution before and after pixel value shifting.}\label{fig:distr:ls8}
\end{figure}

British Columbia was chosen as there were land cover labels available from the NALCMS schema for this region and it has a similar landscape to the Norway region of interest; both are alpine coastal regions that are dominated by forest land cover. The Norway region has LULC labels from the CORINE schema. The availability of land cover labels for the two regions of interest meant that generalized land cover labels could be identified for standardized validation of the domain adaptation map product (i.e., IoU accuracy). The Level I generalized classification scheme is shown in Table~\ref{recoding-table}.
\begin{table}
\tbl{The NALCMS, CORINE and Generalized classification schemes. Only classes which present in the explored areas are presented in the table.}
{\begin{tabular}{lcl} \toprule
 NALCMS & General Land Cover & CORINE\\ \midrule
 
 Temperate or sub-polar needleleaf forest & \multirow{4}{*}{Forest}& Broad-leaved forest \\
 Sub-polar taiga needleleaf forest & & Coniferous forest \\
 \makecell[l]{Temperate or sub-polar broadleaf\\ deciduous forest} & & Mixed forest\\ 
 Mixed forest & & \\ \midrule
 
 Temperate or sub-polar shrubland & \multirow{4}{*}{Grassland}& Transitional woodland shrub \\
 Temperate or sub-polar grassland & & Sparsely vegetated areas \\
 Sub-polar or polar shrubland lichen moss & & Green urban areas\\ 
 Sub-polar or polar grassland lichen moss & & \\ \midrule
 
 Wetland & \multirow{3}{*}{Wetland} & Moors and heathland \\
  & & Inland marshes\\ 
  & & Peat bog\\ \midrule
  
 Cropland & \multirow{4}{*}{Cropland} & Non irrigated arable land \\
  & & Pasture\\ 
  & & Complex cultivation pattern\\
  & &\makecell[l]{Land principally occupied by agriculture \\with significant areas of natural vegetation}\\ \midrule

 Baren lands & \multirow{4}{*}{Barren} & Beaches dunes sands \\
  & & Bare rock\\ 
  & & Burnt areas\\
  & & Mineral extraction site\\ \midrule 
  
 Urban & \multirow{9}{*}{Settlement} & Beaches dunes sands \\
  & & Continuous urban fabric\\ 
  & & Discontinues urban fabric\\
  & & Industrial or commercial units\\ 
  & & \makecell[l]{Road and rail networks and associated\\ lands}\\
  & & Port areas\\ 
  & & Airports\\
  & & Dump site\\ 
  & & Construction site\\
  & & Sport and leisure facilities\\ \midrule 
  
  Water & \multirow{4}{*}{Water} & Intertidal flats\\
  & & Water courses\\ 
  & & Water bodies\\
  & & Sea and ocean\\ \midrule 
  
  Snow and ice  & Snow and glaciers & Glaciers and perpetual snow\\ \bottomrule
\end{tabular}}
\label{recoding-table}
\end{table}
The source dataset used in this work was Landsat 8 imagery acquired in the summer months (June, July, and August) of 2019-2021 years over a region in British Columbia (Canada). The original Level-1 data products of blue, green, red, near-infrared, swir1, and swir2 bands were used in the is analysis. The 6-band composite was cropped to the region of interest (Figure~\ref{fig:roi:bc}). The spatial resolution of all bands was 30m, and the radiometric resolution was 16-bit. Due to the specificity of the Landsat 8 original pixel value distribution, which is shifted to the right by approximately 5000 for each band, it was decided to shift each band value closer to 0, as shown in Figure~\ref{fig:distr:ls8}. The corresponding labels from the NALCMS 2015 with 19 classes was used to define the general LULC for the British Columbia region of interest. The cloud mask was extracted using a quality assessment band, and cloudy areas were masked as {\em nodata}. Finally, the GeoTIFF raster image and corresponding label map were cropped into samples of size 6 $\times$ 512 $\times$ 512 and 1 $\times$ 512 $\times$ 512, respectively, where 6 represents the number of bands in the image file and 1 is a number of bands in the label file. The total number of training files for the source dataset was 1565.

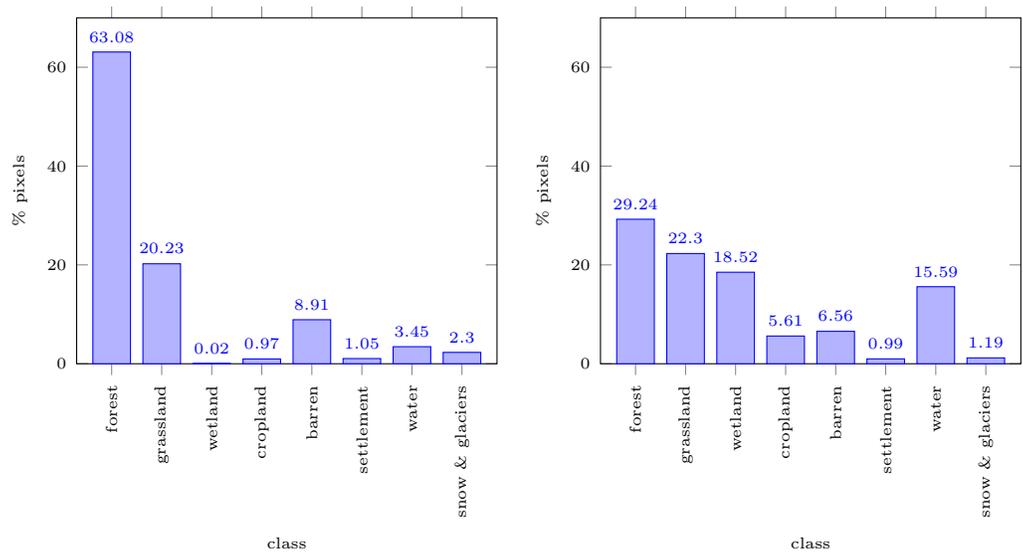
\begin{figure}
\centering
\subfigure[The source dataset label distribution.\label{fig:label:bc}]
{
\begin{tikzpicture}
\tikzstyle{every node}=[font=\tiny]
   \pgfplotsset{
     /pgf/number format/precision=4, % <----
     legend style={font=\footnotesize},
   }
    \begin{axis}[
            ybar,
            bar width=0.5cm,
            width=0.5\textwidth,
            symbolic x coords={forest,grassland,wetland,cropland,barren,settlement,water,snow \& glaciers},
            xtick=data,
            xlabel = {class},
            nodes near coords,
            nodes near coords align={vertical},
            ymin=0,ymax=70,
            ylabel={\% pixels},
            y tick label style={font=\tiny},
            xticklabel style={rotate=90},
            x tick label style={font=\tiny}
        ]
        \addplot table[x=class,y=ls]{\mydata};
    \end{axis}
\end{tikzpicture}
}
\subfigure[The target dataset label distribution.\label{fig:label:nw}]
{
\begin{tikzpicture}
\tikzstyle{every node}=[font=\tiny]
   \pgfplotsset{
     /pgf/number format/precision=4, % <----
     legend style={font=\footnotesize},
   }
    \begin{axis}[
            ybar,
            bar width=0.5cm,
            width=0.5\textwidth,
            symbolic x coords={forest,grassland,wetland,cropland,barren,settlement,water,snow \& glaciers},
            xtick=data,
            xlabel = {class},
            nodes near coords,
            nodes near coords align={vertical},
            ymin=0,ymax=70,
            ylabel={\% pixels},
            y tick label style={font=\tiny},
            x tick label style={font=\tiny},
            xticklabel style={rotate=90}
        ]
        \addplot table[x=class,y=sent]{\mydata};
    \end{axis}
\end{tikzpicture}
}
\caption{Label distribution in the source and target datasets.}\label{fig:labels}
\end{figure}

The target dataset was represented by Sentinel-2 imagery (L1C preprocessing level), taken in June, July, and August of 2019-2021 over the Norway area (Figure~\ref{fig:roi:nw}). The original tiles were merged, and the same 6-band composite was created as in the source dataset and cropped to the Norway region of interest. The spatial resolution of all Sentinel-2 bands was 20m, and the radiometric resolution was 16-bit. The CORINE 2020 dataset with 31 classes in Norway was used to define the general LULC labels. The merged Sentinel-2 composition and corresponding label raster were cropped into the tiles, the same as for the source dataset. The total number of training files for the target dataset was 2461. The final LULC class label distribution for both datasets is depicted in Figure~\ref{fig:labels}.
\pgfplotstableread[row sep=\\,col sep=&]{
    class & ls & sent\\
    forest	& 63.08  & 29.24 \\
    grassland	& 20.23 & 22.30 \\
    wetland   & 0.02 & 18.52 \\
    cropland   & 0.97 & 5.61 \\
    barren   & 8.91  & 6.56 \\
    settlement   & 1.05  & 0.99 \\
    water   & 3.45  & 15.59 \\
    snow \& glaciers & 2.30  & 1.19 \\
}\mydata
    
%%British Columbia was chosen as the NALCMS labels are available for this region and similar land covers such as alpine, ocean and forest land cover found in the Norway region of interest. Norway region has the CORINE LULC schema available for testing the harmonization validation. {\em Here, I would like to see a paragraph from Jami with a discussion of validation methods and their uniqueness}

The adaptation model was run with the same hyperparameters as in \citet{sokolov} except the number of the input and output bands was increased from 4 to 6. The image values from both domains were rescaled from 16-bit format to the values on the range [-1, 1]. After the domain adaptation model was trained, the inferences were generated by back scaling from the float point values in the range [-1, 1] to the 16-bit integer format; thus, all source domain images are transferred to the style of the target domain. Then, a segmentation model was used to evaluate the quality of the translated images. It was represented by the DeepLab v3 framework with a modified number of input channels equal to 6. As an optimizer, the Adam method was chosen with the initial learning rate and weight decay equal to 1 $\times$ $10^{-4}$ and 5 $\times$ $10^{-4}$, respectively. During the segmentation model training, the learning rate was adjusted using the polynomial decay method with a power of 0.9. The model was trained with eight images in a batch and committed 90,000 steps. Random rotation and flipping was applied to the training samples in order to augment the training dataset. For the validation, the entire target dataset was used. Before training the segmentation model, each mean band value was calculated for all datasets (training and validation) and then subtracted from each training sample.

\section{Results}
The performance of the semantic segmentation model trained with the adapted images was compared to the performance of the baseline model. The architecture of the baseline model is the same as the segmentation model trained on the adapted images. The baseline segmentation model was trained only on the source imagery (Landsat 8) and validated on the target domain (Sentinel-2).

\begin{figure}
	\centering
	\includegraphics[width=11cm]{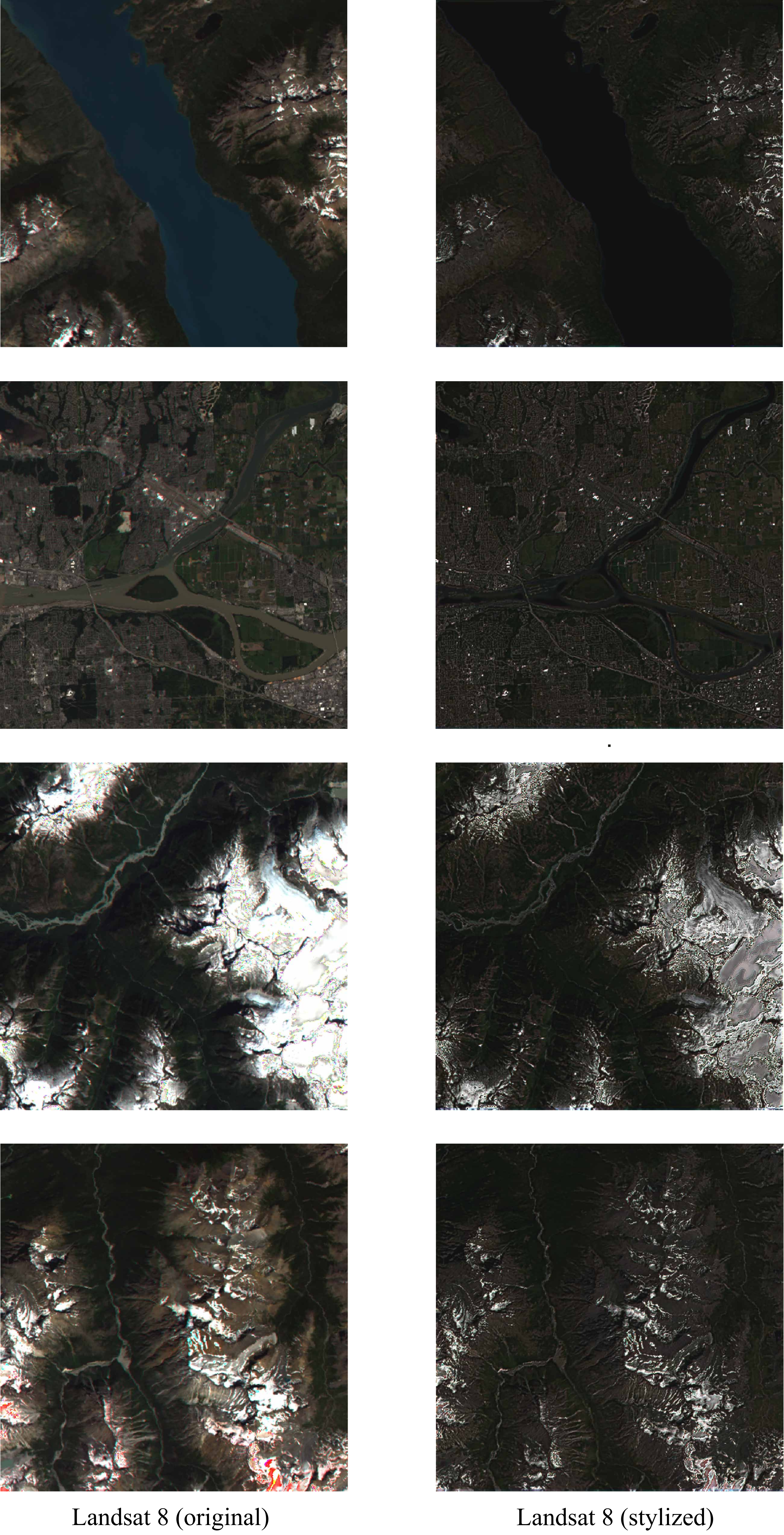}
	\caption{Examples showing the result of the style-transferring. The stylized images look sharper because spatial resolution of the target domain is higher. Also, the tone of the objects was changed according to the target domain style.}
	\label{fig:results:da}
\end{figure}

From Table~\ref{results-table}, it can be seen that the segmentation model trained with the domain-adapted samples shows an increase in the mean intersection-over-union (mIoU) performance of almost 77\% compared to the baseline. Also, some semantic classes, such as cropland and settlement, became more {\em visible} after the domain adaptation compared to the baseline model. The IoU performance for the wetland class is zero for both models due to poor representation in the source domain (0.02\%), and wetland areas are generally hard to discriminate \citep{j1, j2, j3}. In addition, the NALCMS label map is dated 2015, whereas Landsat 8 images were acquired in 2019-2021. Due to last-year dry summers in Canada, the labels are highly likely to mismatch the wetland areas in the images \citep{heat}.
\begin{table}
\tbl{The segmentation models'  performance (IoU).}
{\begin{tabular}{lccccccccc} \toprule
 Method & mIoU & Forest & Grassland & Wetland & Cropland & Barren & Settlement & Water &\makecell[c]{Snow \\\& glaciers} \\ \midrule
 Baseline & 18.55 & 35.19 & 22.98 & 0 & 0 & 5.80 & 0 & 72.46 & 11.98\\ \midrule
 HRSemI2I & 32.83 & 53.82 & 22.46 & 0 & 19.92 & 22.82 & 33.33 & 81.51 & 28.80 \\ \bottomrule
\end{tabular}}
\label{results-table}
\end{table}
The results of the style-transferring from the Sentinel-2 domain to the Landsat 8 are presented in Figure~\ref{fig:results:da}. It is important to notice that the HRSemI2I model preserved the semantic consistency during the style-transferring operation; thus, all Landsat 8 images were kept semantically unchanged, which is crucial for the segmentation task in remote sensing. Furthermore, we can see that the style-translated images look sharper than their originals as the model compares the source (30 m) and target (20 m) domain images spatial resolution. Hence, the HRSemI2I model transfers the radiometric style of the target domain and its spatial resolution.

\begin{figure}
	\centering
	\includegraphics[width=12cm]{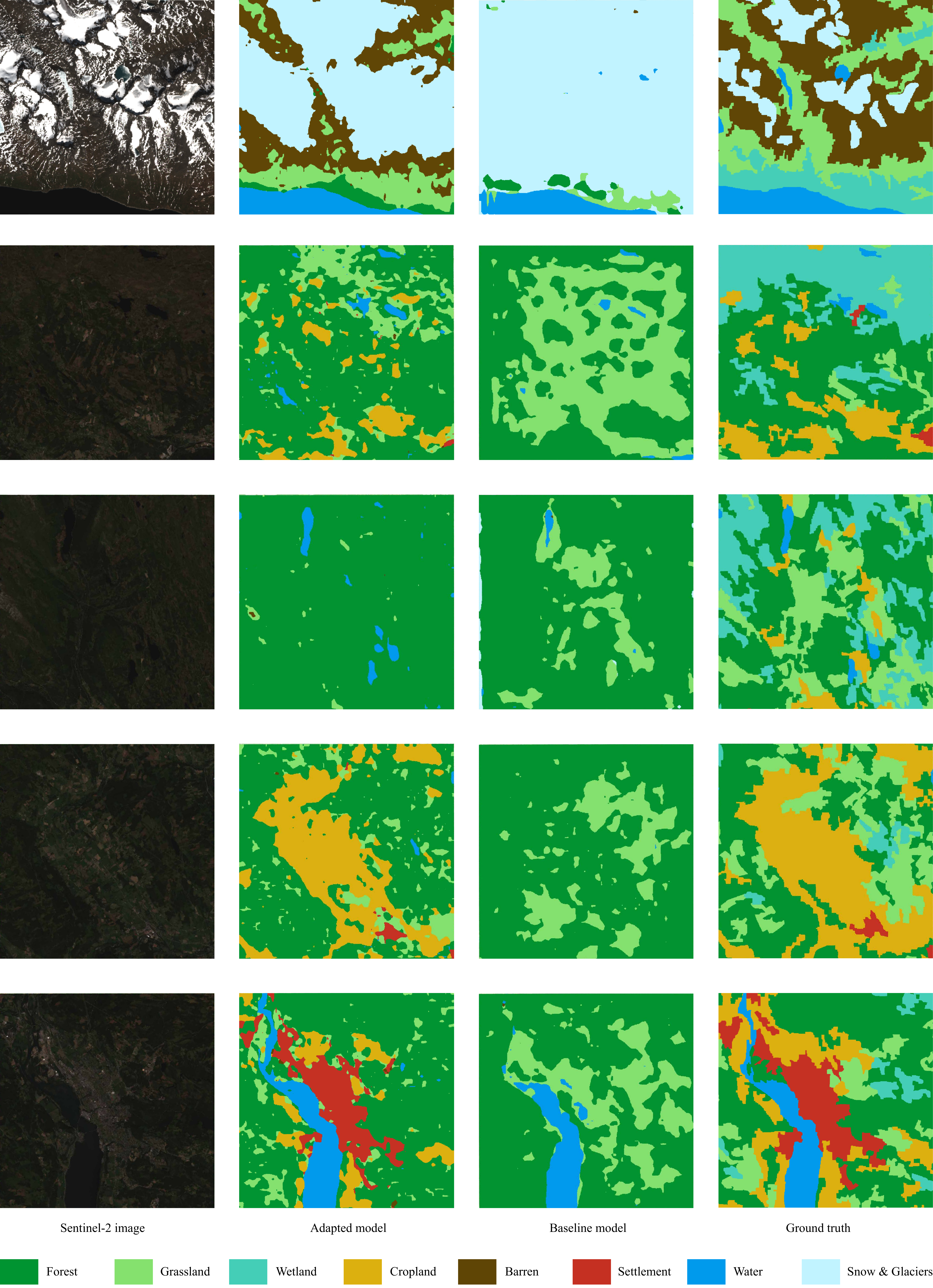}
	\caption{Examples showing the results of the semantic segmentation.}
	\label{fig:results:segm}
\end{figure}

Visual results of the semantic segmentation of the target domain (Sentinel-2) are presented in Figure~\ref{fig:results:segm}. This work is a first step towards sensor-to-sensor domain adaptions for map automation. The examples show how the adapted model generates more accurate labels than the baseline model. For instance, the urban settlement class is completely invisible for the baseline model, whereas the adapted model detects it. We can observe the same in the second-from-the-bottom-row example, where the cropland class was noticeably better detected by the adapted model while misclassified by the model with no adaptation. As for the barren and snow classes, the adapted model also shows a significant advantage which can be seen in the upper-most-row example. The baseline model could not classify barren structures and mistakenly classified everything as snow and glaciers class. The second and third-from-the-top-row examples show the low performance for both models regarding the wetland class. However, the adapted model is still better in terms of detecting waterbodies and cropland areas.

\section{Conclusion}
It is hard to overestimate the meaning of actual and easily accessible satellite data today. It is used by private companies and governments all over the world. Such satellite constellations as Landsat and Sentinel are prominent representatives of free and frequent satellite data providers. However, the pace of satellite imagery generation and their labelling is significantly different, which decreases their usability. Also, the discrepancy in sensors’ characteristics plays a substantial role, making the inferences of the classic segmentation models useless. The solution to this problem could be domain adaptation.

In this article, the example of domain adaptation between Landsat and Sentinel satellite images was considered using the HRSemI2I model. The results were compared with the model without adaptation. From the results, the DA significantly improves the segmentation performance both overall and per-class in this initial exploration. We recognized that the accuracy results do not meet the expected accuracies for remote sensing, however, the results suggests that further development in this direction makes sense.

For future development, the following improvements could be made. First, more actual and accurate label maps could bring even higher segmentation accuracy. Considered in this article, label schemes (NALCMS 2015 and CORINE) provide only a rough estimation of the class boundaries. Second, the wetlands class should be detected separately since it is highly dependable on current weather conditions. Preferably, other methods of detection should be applied, independent of the ground truth labels (e.g. thresholding, etc.).

\section*{Disclosure statement}
The authors have no conflicts of interest to declare. All co-authors have seen and agree with the contents of the manuscript and there is no financial interest to report. We certify that the submission is original work and is not under review at any other publication.

\section*{Funding}
DIKU - Norwegian Agency for International Cooperation and Quality Enhancement in Higher Education\\
PNA-2019/10090 – NOK \$300,000\\
Map production using advanced techniques in neural networks and remote sensing\\
Principal Investigator: Rune Strand Ødegård

\end{document}